\documentclass[letterpaper, 10 pt, conference]{ieeeconf}
\IEEEoverridecommandlockouts                              
\newcommand{\arxiv}[2]{#1}

\usepackage{times}
\usepackage{epsfig}
\usepackage{graphicx}
\usepackage{amssymb}

\usepackage{graphicx}
\usepackage{epsfig}
\usepackage{times}
\usepackage{mathptmx}
\usepackage{cite}
\usepackage{color}
\usepackage{times}

\definecolor{red}{rgb}{1,0,0}
\definecolor{green}{rgb}{0,1,0}
\definecolor{blue}{rgb}{0,0,1}
\definecolor{violet}{rgb}{1,0,1}
\definecolor{cyan}{cmyk}{1,0,0,0}
\definecolor{magenta}{cmyk}{0,1,0,0}
\definecolor{yellow}{cmyk}{0,0,1,0}

\definecolor{white}{rgb}{1,1,1}

\newcommand{\CO}[1]{}

\newcommand{\CommentOut}[1]{}

\begin{document}

\newcommand{\FIG}[3]{
\begin{minipage}[b]{#1cm}
\begin{center}
\includegraphics[width=#1cm]{#2}\\
{\scriptsize #3}
\end{center}
\end{minipage}
}

\newcommand{\FIGU}[3]{
\begin{minipage}[b]{#1cm}
\begin{center}
\includegraphics[width=#1cm,angle=180]{#2}\\
{\scriptsize #3}
\end{center}
\end{minipage}
}

\newcommand{\FIGm}[3]{
\begin{minipage}[b]{#1cm}
\begin{center}
\includegraphics[width=#1cm]{#2}\\
{\scriptsize #3}
\end{center}
\end{minipage}
}

\newcommand{\FIGR}[3]{
\begin{minipage}[b]{#1cm}
\begin{center}
\includegraphics[angle=-90,width=#1cm]{#2}
\\
{\scriptsize #3}
\vspace*{1mm}
\end{center}
\end{minipage}
}

\newcommand{\FIGRpng}[5]{
\begin{minipage}[b]{#1cm}
\begin{center}
\includegraphics[bb=0 0 #4 #5, angle=-90,clip,width=#1cm]{#2}\vspace*{1mm}
\\
{\scriptsize #3}
\vspace*{1mm}
\end{center}
\end{minipage}
}

\newcommand{\FIGpng}[5]{
\begin{minipage}[b]{#1cm}
\begin{center}
\includegraphics[bb=0 0 #4 #5, clip, width=#1cm]{#2}\vspace*{-1mm}\\
{\scriptsize #3}
\vspace*{1mm}
\end{center}
\end{minipage}
}

\newcommand{\FIGtpng}[5]{
\begin{minipage}[t]{#1cm}
\begin{center}
\includegraphics[bb=0 0 #4 #5, clip,width=#1cm]{#2}\vspace*{1mm}
\\
{\scriptsize #3}
\vspace*{1mm}
\end{center}
\end{minipage}
}

\newcommand{\FIGRt}[3]{
\begin{minipage}[t]{#1cm}
\begin{center}
\includegraphics[angle=-90,clip,width=#1cm]{#2}\vspace*{1mm}
\\
{\scriptsize #3}
\vspace*{1mm}
\end{center}
\end{minipage}
}

\newcommand{\FIGRm}[3]{
\begin{minipage}[b]{#1cm}
\begin{center}
\includegraphics[angle=-90,clip,width=#1cm]{#2}\vspace*{0mm}
\\
{\scriptsize #3}
\vspace*{1mm}
\end{center}
\end{minipage}
}

\newcommand{\FIGC}[5]{
\begin{minipage}[b]{#1cm}
\begin{center}
\includegraphics[width=#2cm,height=#3cm]{#4}~$\Longrightarrow$\vspace*{0mm}
\\
{\scriptsize #5}
\vspace*{8mm}
\end{center}
\end{minipage}
}

\newcommand{\FIGf}[3]{
\begin{minipage}[b]{#1cm}
\begin{center}
\fbox{\includegraphics[width=#1cm]{#2}}\vspace*{0.5mm}\\
{\scriptsize #3}
\end{center}
\end{minipage}
}


\newcommand{\acprPaperID}{25}





\newcommand{\noeditage}[1]{#1} \newcommand{\editage}[1]{}


\title{\bf \Large
Deep SIMBAD:\\
Active Landmark-based Self-localization Using Ranking -based Scene Descriptor 
}

\author{Tanaka Kanji
\thanks{Our work has been supported in part by 
JSPS KAKENHI 
Grant-in-Aid 
for Scientific Research (C) 17K00361 
and 20K12008.}
\thanks{The authors are with Graduate School of Engineering, University of Fukui, Japan. 
{\tt\small tnkknj@u-fukui.ac.jp}}}

\newcommand{\figA}{
\begin{figure}[b]
\begin{flushright}
\vspace*{-8mm}
\hspace*{5mm}\FIG{8.5}{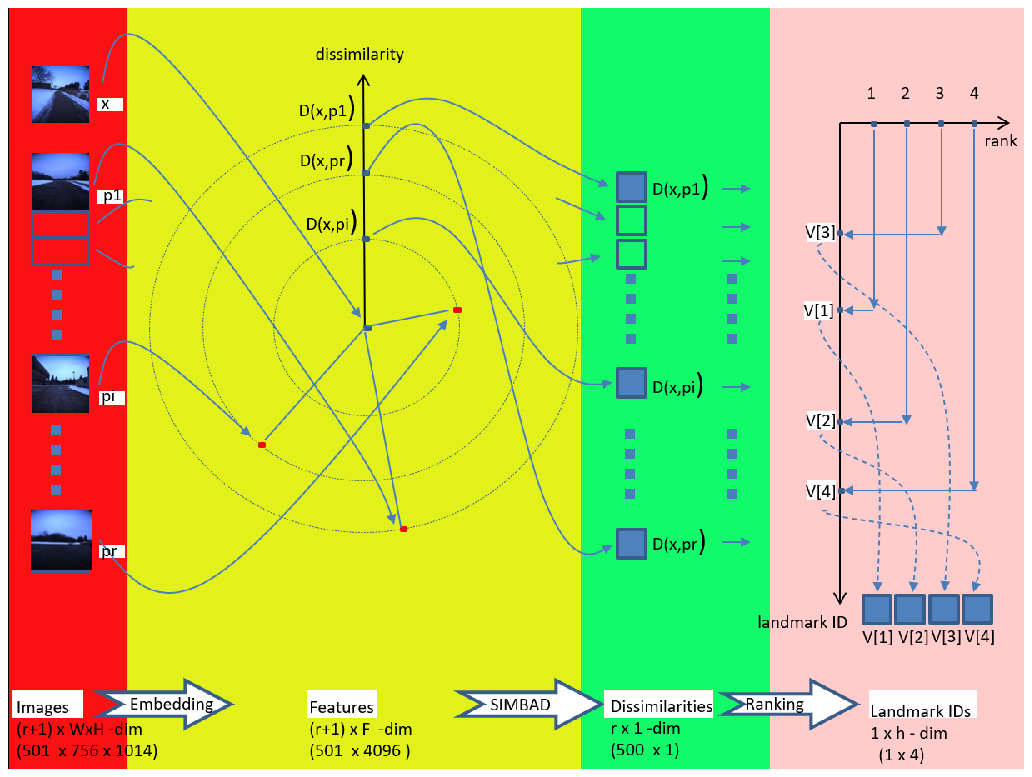}{}
\caption{Scene description module. First, the input query/landmark images are translated into vectorial features. Subsequently, the query feature is described by its dissimilarities from the landmark features. Next, the dissimilarity values are ranked to obtain a ranked list of top-$h$ most similar landmarks. Finally, the proposed landmark ranking-based scene descriptor is obtained.}\label{fig:A}
\end{flushright}
\end{figure}
}

\newcommand{\figB}{
\begin{figure}[b]
\vspace*{-8mm}
\FIG{8.5}{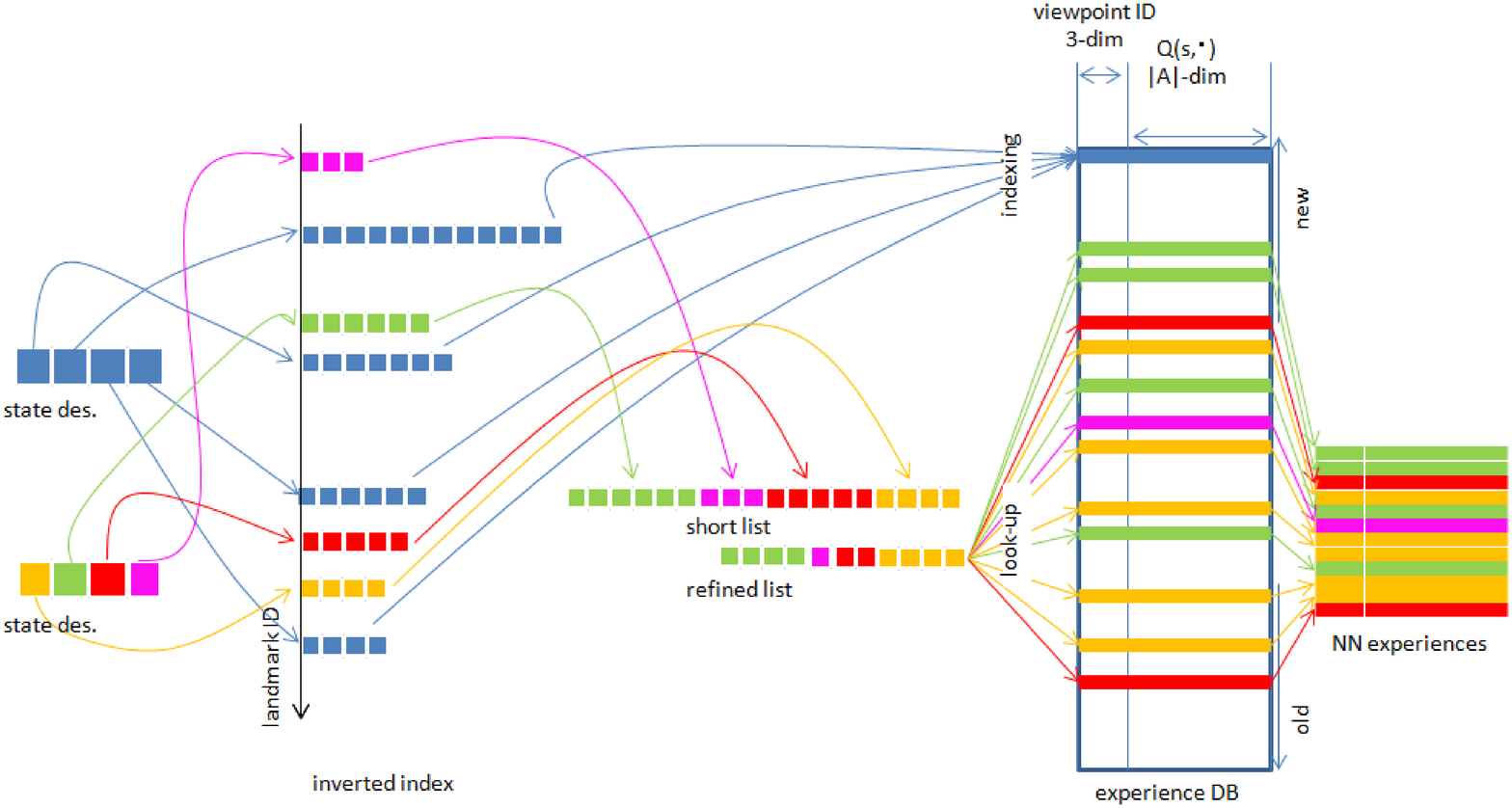}{}\vspace*{-5mm}%
\caption{Nearest neighbor scheme. (1) Offline mapping stage: First, a map image is translated to a landmark ranking-based scene descriptor. Subsequently, the image ID is inserted into the inverted index entries that have common landmark IDs. (2) Online self-localization stage: First, a live image is translated into a landmark ranking-based scene descriptor. Next, a shortlist is obtained from the inverted index entries that have common landmark IDs, and then refined. Finally, the viewpoints or Q-values associated with the nearest neighbor states are obtained from the experience database, and then returned.}\label{fig:B}
\end{figure}
}

\newcommand{\figC}{
\begin{figure}[b]
\begin{center}
\vspace*{-8mm}
\FIG{8}{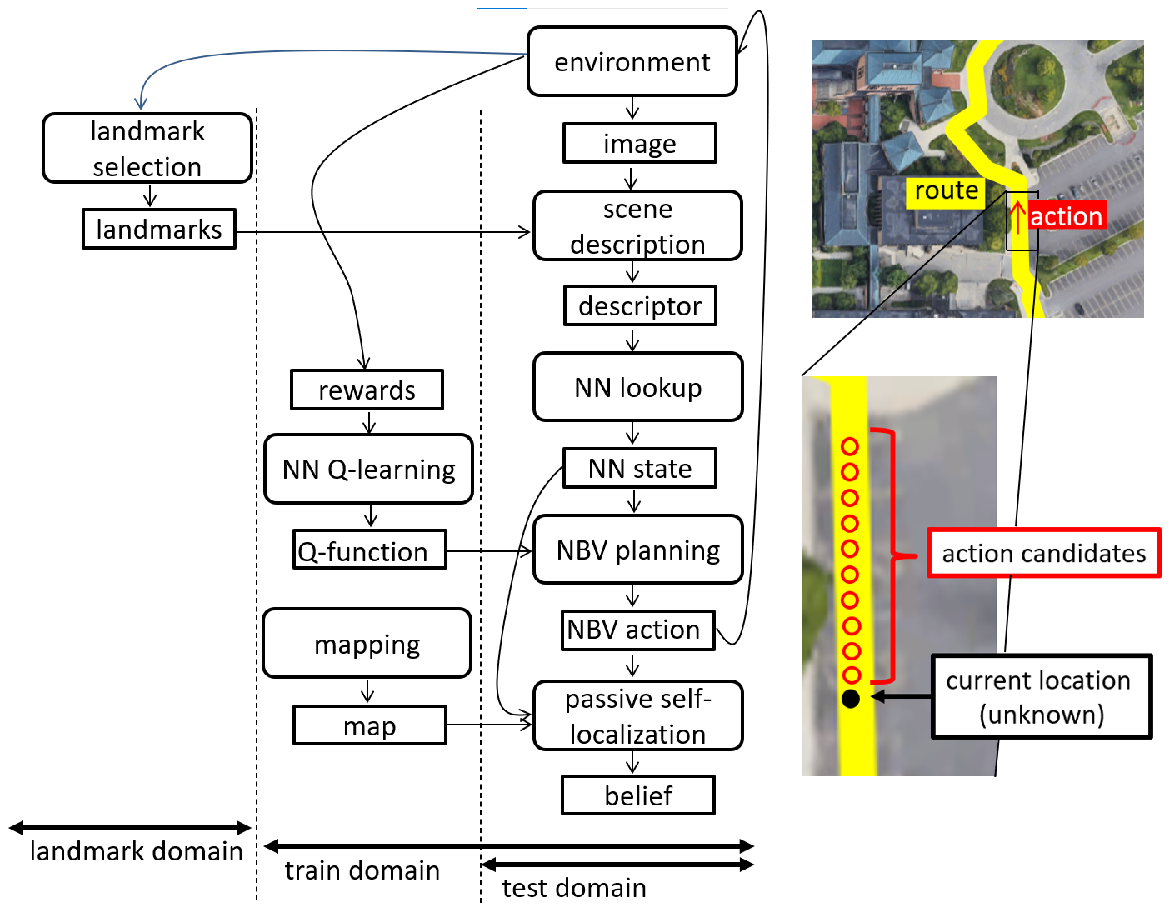}{}
\caption{
Active self-localization system. Online processing comprised three main stages: scene description, passive self-localization, and NBV planning. Offline processing comprised landmark selection, mapping, and NN Q-learning. Details of the scene description module are provided in Fig. \ref{fig:A}. Details of the NN lookup module are provided in Fig. \ref{fig:B}.
}\label{fig:C}
\end{center}
\end{figure}
}

\newcommand{\figD}{
\begin{figure}[b]
\begin{center}
\vspace*{-8mm}
\FIG{8}{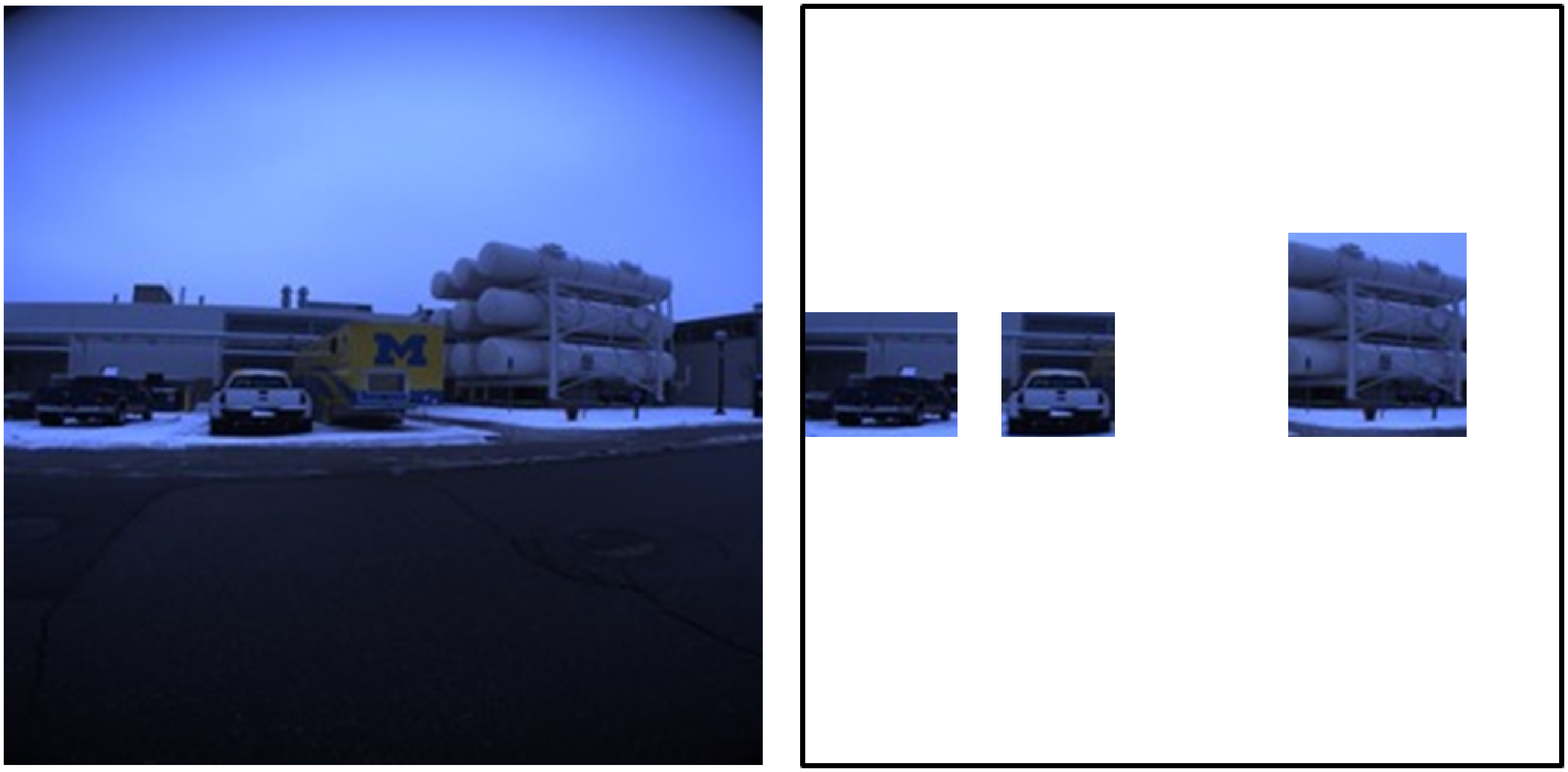}{}\vspace*{-3mm}%
\caption{Motivation. Unlike existing approaches that maintain the model of an entire environment (left panel), we aim to maintain only a small fraction (i.e., landmark regions) of the robot workspace (right panel). Hence, the per-domain cost for retraining (change detection, map updating) is inherently low. However, self-localization with such spatially sparse landmarks can be an ill-posed problem for a passive observer, as many viewpoints may not provide an effective landmark view. Therefore, we considered an active observer and presented an active landmark-based self-localization framework.
}\label{fig:D}
\end{center}
\end{figure}
}

\newcommand{\figE}{
\begin{figure}[b]
\vspace*{-5mm}
\begin{minipage}[b]{8cm}
\FIGpng{8}{../03-1333216490407637-11-1353174220475252.png}{}{2926}{256}
\FIGpng{8}{../01-1327254351290835-11-1353174227874337.png}{}{2926}{256}
\FIGpng{8}{../03-1333220969150702-01-1327250156410637.png}{}{2926}{256}
\FIGpng{8}{../01-1327250119606322-11-1353174165281978.png}{}{2926}{256}
\FIGpng{8}{../11-1353176746162084-03-1333216088176341.png}{}{2926}{256}
\FIGpng{8}{../01-1327253696815977-08-1344081644194009.png}{}{2926}{256}
\FIGpng{8}{../11-1353178003005860-03-1333216107577860.png}{}{2926}{256}
\FIGpng{8}{../01-1327252089232432-08-1344081705800713.png}{}{2926}{256}
\FIGpng{8}{../03-1333217984521927-01-1327250152610195.png}{}{2926}{256}
\FIGpng{8}{../11-1353176867147012-03-1333216094976881.png}{}{2926}{256}
\vspace*{-7mm}
\end{minipage}
\caption{Examples of similarity-based scene description. First column: input image;
$(i+1)$-th column ($i\ge 1$): $i$-th most similar landmark.}\label{fig:E}
\end{figure}
}

\newcommand{\figH}{
\begin{figure}[b]
\vspace*{-5mm}
\FIG{8}{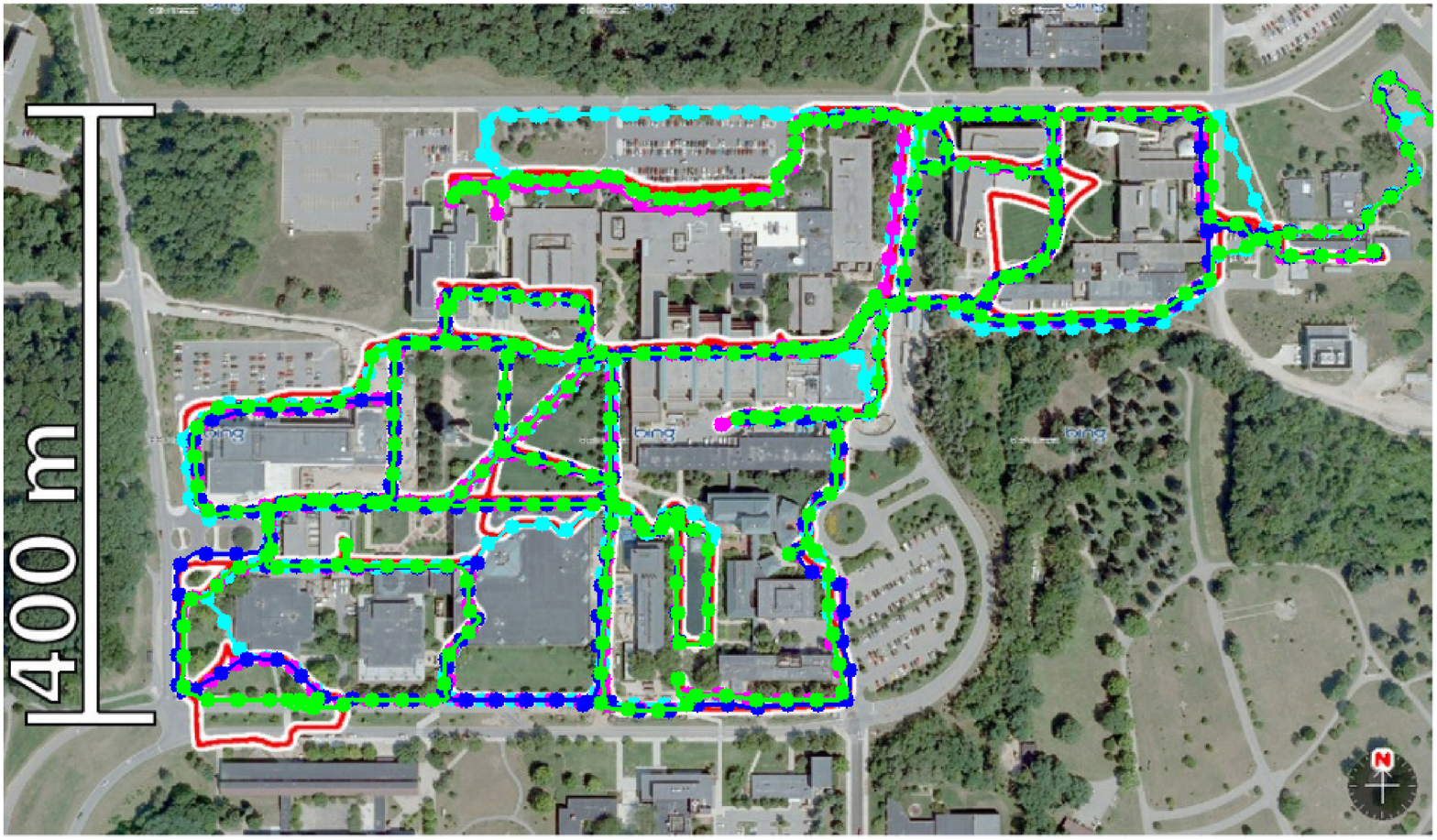}{}\vspace*{-1mm}%
\caption{Experimental environments.
The trajectories of the four datasets,
``2012/1/22", ``2012/3/31", ``2012/8/4", and ``2012/11/17",
used in our experiments are visualized 
in green, purple, blue, and light-blue
curves, respectively.
}\label{fig:H}
\end{figure}
}

\newcommand{\figFJK}{
\begin{figure}[b]
\vspace*{-5mm}
\FIGR{3}{../figF.eps}{(a)}\hspace*{-5mm}%
\FIGR{3}{../figJ.eps}{(b)}\hspace*{-5mm}%
\FIGR{3}{../figK.eps}{(c)}\vspace*{-5mm}%
\caption{ANR of single-view VPR.}\label{fig:FJK}
\end{figure}
}

\arxiv{

\renewcommand{\figFJK}{
\begin{figure}[t]
\FIGR{8}{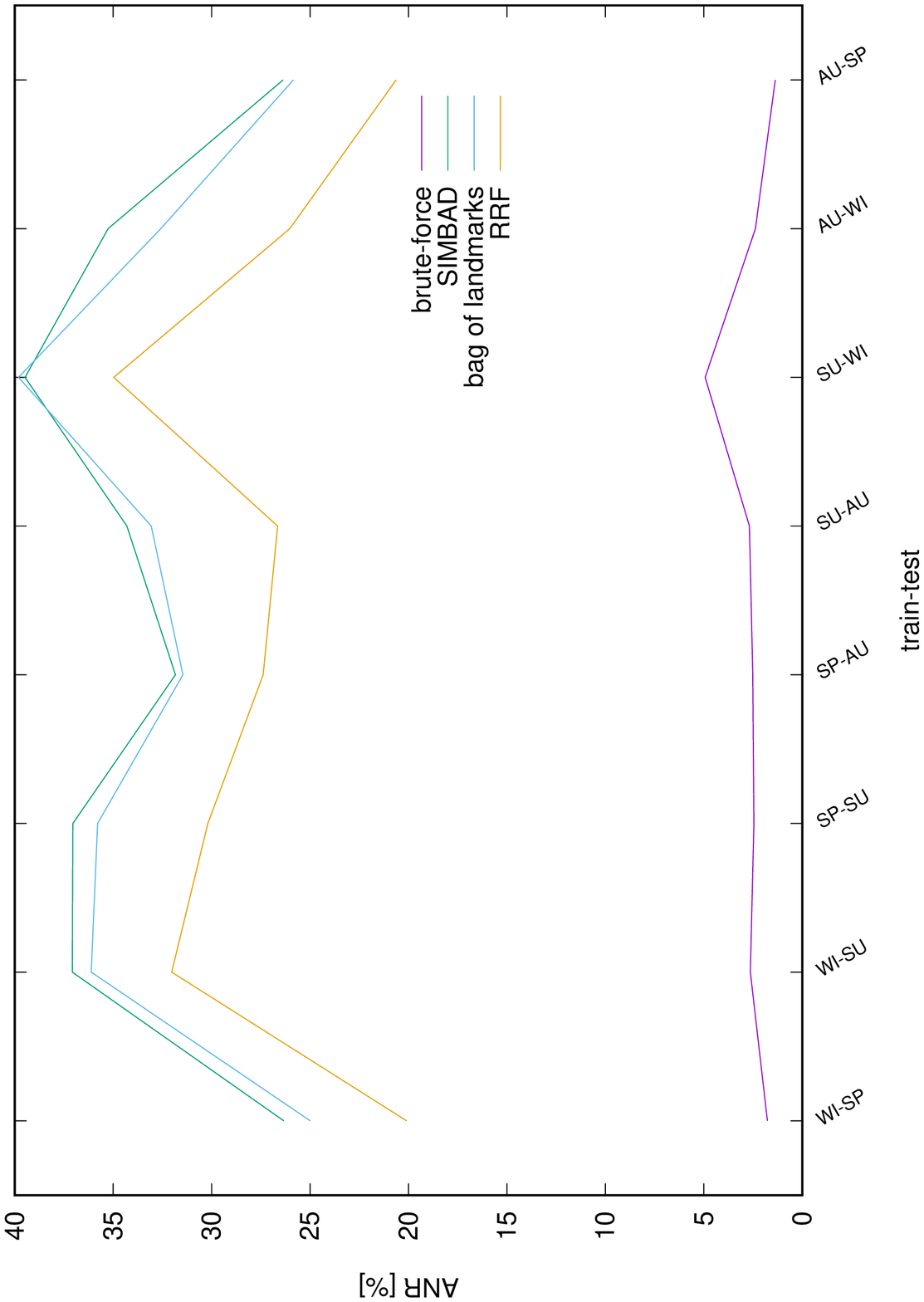}{(a)}
\FIGR{8}{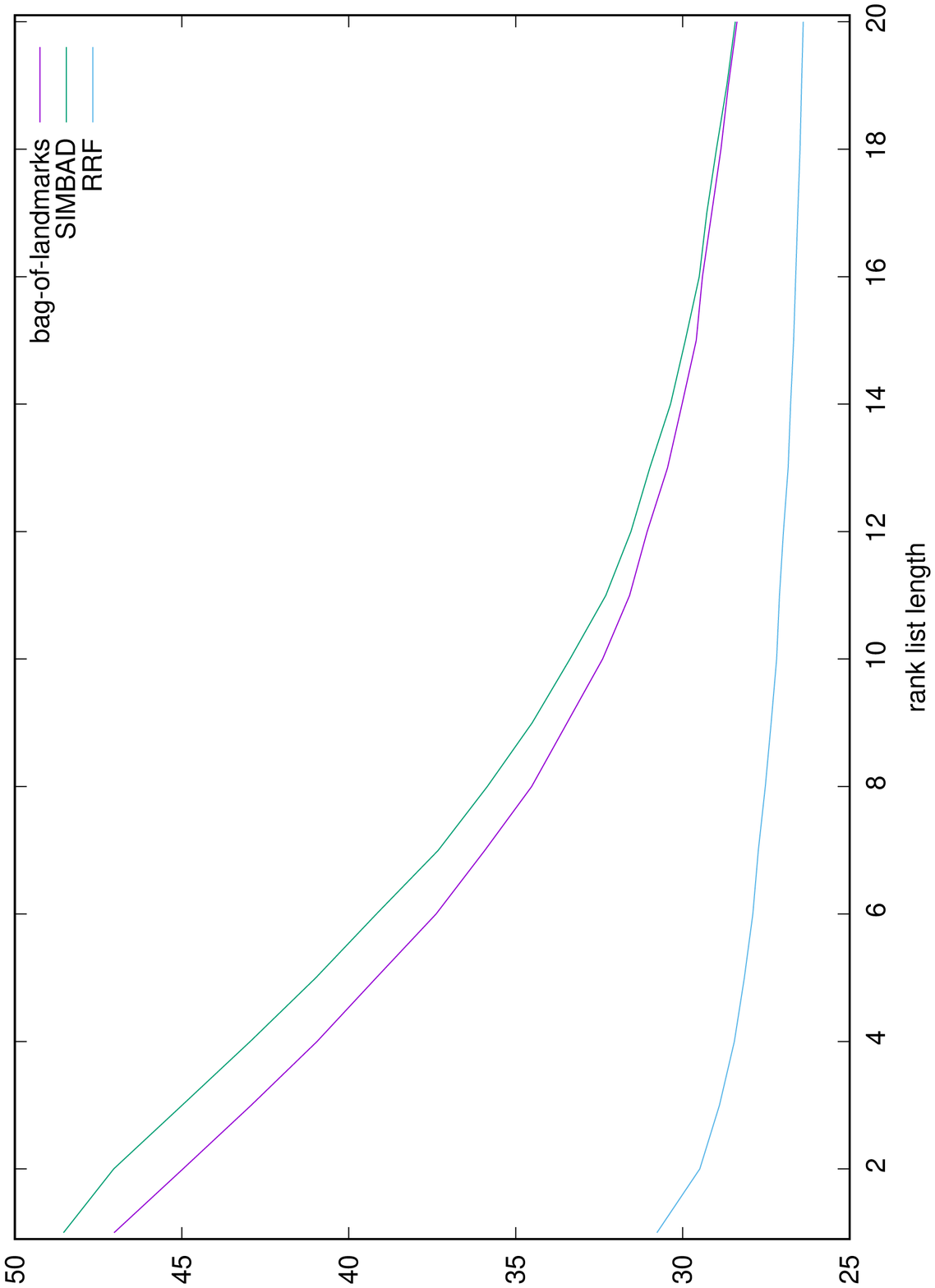}{(b)}
\vspace*{-5mm}
\caption{ANR performance of VPR.}\label{fig:FJK}
\vspace*{-5mm}
\end{figure}
}

}{}

\newcommand{\figG}{
\begin{figure}[b]
\vspace*{-8mm}
\FIGR{8}{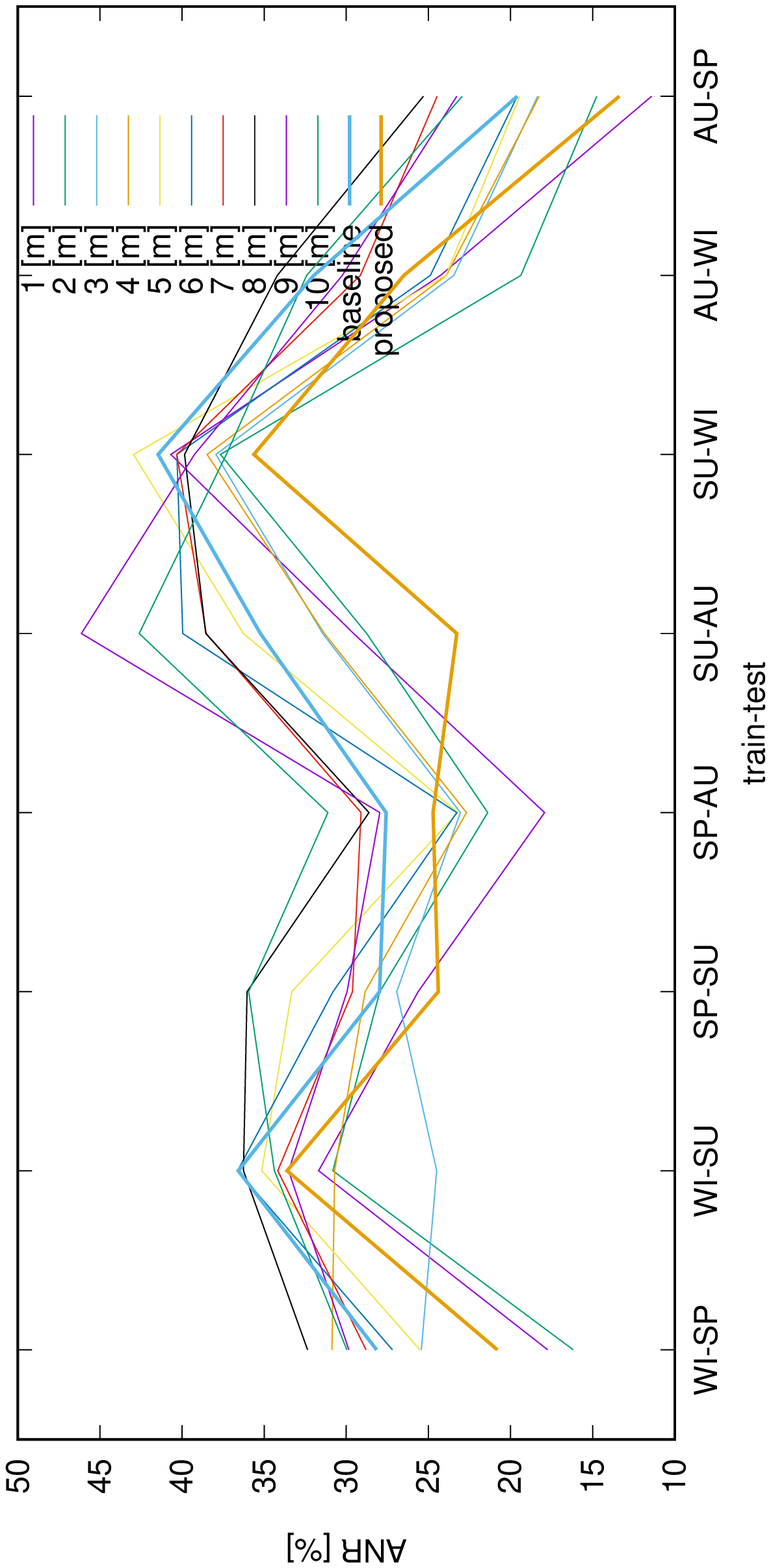}{}\vspace*{-5mm}%
\caption{Performance results for active self-localization.}\label{fig:G}
\end{figure}
}

\newcommand{\fortabA}[1]{
\scriptsize
Top-(#1\%) ANR [\%]\\
\begin{tabular}{r|r|r|r|r|r|r|r|r|}
\hline
\input tab_#1.txt
\end{tabular}\vspace*{1mm}\\
}

\newcommand{\tabA}{
\begin{table*}[b]
\begin{center}
\caption{ANR performance.}\label{tab:A}

\scriptsize
Top-(10\%) ANR [\%]\\
\begin{tabular}{r|r|r|r|r|r|r|r|r|}
\hline
\input tab_10.txt
\end{tabular}\vspace*{1mm}\\

\scriptsize
Top-(20\%) ANR [\%]\\
\begin{tabular}{r|r|r|r|r|r|r|r|r|}
\hline
\input tab_20.txt
\end{tabular}\vspace*{1mm}\\

\scriptsize
Top-(30\%) ANR [\%]\\
\begin{tabular}{r|r|r|r|r|r|r|r|r|}
\hline
\input tab_30.txt
\end{tabular}\vspace*{1mm}\\

\scriptsize
Top-(40\%) ANR [\%]\\
\begin{tabular}{r|r|r|r|r|r|r|r|r|}
\hline
\input tab_40.txt
\end{tabular}\vspace*{1mm}\\

\end{center}
\end{table*}
}

\newcommand{\fortabB}[1]{
\scriptsize
Top-(#1\%) ANR [\%]\\
\begin{tabular}{r|r|r|r|r|}
\hline
\input tabB_#1.txt
\end{tabular}\vspace*{1mm}\\
}

\newcommand{\tabB}{
\begin{table}
\begin{center}
\caption{ANR performance.}\label{tab:A}

\scriptsize
Top-(10\%) ANR [\%]\\
\begin{tabular}{r|r|r|r|r|}
\hline
\input tabB_10.txt
\end{tabular}\vspace*{1mm}\\

\scriptsize
Top-(20\%) ANR [\%]\\
\begin{tabular}{r|r|r|r|r|}
\hline
\input tabB_20.txt
\end{tabular}\vspace*{1mm}\\

\scriptsize
Top-(30\%) ANR [\%]\\
\begin{tabular}{r|r|r|r|r|}
\hline
\input tabB_30.txt
\end{tabular}\vspace*{1mm}\\

\scriptsize
Top-(40\%) ANR [\%]\\
\begin{tabular}{r|r|r|r|r|}
\hline
\input tabB_40.txt
\end{tabular}\vspace*{1mm}\\

\end{center}
\end{table}
}

\maketitle

\begin{abstract}
Landmark-based robot self-localization has recently garnered interest as a highly-compressive domain-invariant approach for performing visual place recognition (VPR) across domains (e.g., time of day, weather, and season). However, landmark-based self-localization can be an ill-posed problem for a passive observer (e.g., manual robot control), as many viewpoints may not provide an effective landmark view. In this study, we consider an active self-localization task by an active observer and present a novel reinforcement learning (RL)-based next-best-view (NBV) planner. Our contributions are as follows. (1) SIMBAD-based VPR: We formulate the problem of landmark-based compact scene description as SIMBAD (similarity-based pattern recognition) and further present its deep learning extension. (2) VPR-to-NBV knowledge transfer: We address the challenge of RL under uncertainty (i.e., active self-localization) by transferring the state recognition ability of VPR to the NBV. (3) NNQL-based NBV: We regard the available VPR as the experience database by adapting nearest-neighbor approximation of Q-learning (NNQL). The result shows an extremely compact data structure that compresses both the VPR and NBV into a single incremental inverted index. Experiments using the public NCLT dataset validated the effectiveness of the proposed approach.
\end{abstract}

\section{%
Introduction
}

Landmark-based robot self-localization has recently garnered interest as a highly-compressive domain-invariant approach for performing visual place recognition (VPR) across domains (e.g., time of day, weather, season). In long-term cross-domain navigation \cite{cs1, cs3, ref28}, a robot vision must be able to recognize its location (i.e., self-localization) and the main objects \cite{torralba2003context} (i.e., landmarks) in the scene. This landmark-based self-localization problem presents two unique challenges. (1) Landmark selection: In the offline training stage, the robot must learn the main landmarks that represent the robot workspace in either a self-supervised or unsupervised manner \cite{topo3}. (2) Next best view (NBV): In the online self-localization stage, the robot must determine the NBVs to re-identify as many spatially sparse landmarks as possible \cite{deepactivelocalization}. This study focuses on the NBV problem.

In landmark-based self-localization, 
the only available feature (Fig. \ref{fig:D}) in each map/live scene $x$ is landmark 
$p_i (i\in [1, r])$ observed at signal strength $d(x, p_i)$ (i.e., landmark ID + intensity).
This is in contrast to many existing self-localization frameworks that assume the availability of vectorial features for each map image.
To address this issue, we formulate the landmark-based self-localization as similarity-based pattern recognition (SIMBAD) \cite{feragen2015similarity}  and present its deep-learning extension. In pattern recognition, SIMBAD is a highly-compressive method to describe an object as its dissimilarities from 
$r$ prototypes (e.g., $r=500$), which is particularly effective when traditional vectorial descriptions of objects are difficult to obtain or inefficient for learning. 
We adopt SIMBAD because of the following two reasons: 
(1)
First, the landmark selection problem is analogous to and can be
informed by the well-investigated problem of prototype selection or optimization \cite{gurumoorthy2021spot}. 
(2)
Second, many recent deep learning techniques can effectively measure the (dis)similarity between an object and prototypes 
\cite{li2019siamrpn++}, 
instead of directly describing an object. 
However, the deep learning extension of SIMBAD has not yet been sufficiently investigated, and hence is the main focus of our current study.

\noeditage{
\figD
}

Most existing self-localization techniques reported hitherto assume a passive observer (e.g., manual robot control) and do not consider the issue of viewpoint planning or observer control. 
However, landmark-based self-localization can be an ill-posed problem for a passive observer, as many viewpoints may not provide an effective landmark view. Therefore, we aim to develop an active observer that can adapt its viewpoints, thereby avoiding non-salient scenes that provide no effective landmark view, or moving efficiently toward locations that are the most informative, to reduce sensing/computation costs. This is associated closely with the NBV problem investigated in machine vision studies \cite{mendoza2020supervised}. However, in our cross-domain scenario, the NBV planner is trained and tested in different domains. The cost for retraining such an NBV planner that does not consider domain shifts is high in cross-domain scenarios, and we intend to address this issue in the current study.

In this study, we investigated an active landmark-based self-localization framework from a novel perspective of SIMBAD. Our primary contributions are threefold: (1) SIMBAD-based VPR: First, we present a landmark-based compact scene descriptor by introducing a deep-learning extension of SIMBAD. Our strategy is to describe each map/live scene $x$ with dissimilarities 
$\{d(x, p_i)\}$ 
from the prototypes (i.e., landmarks) 
$\{p_i\}_{i=1}^r$ 
and further compress it into a length 
$h$ $(\ll r)$ list of ranked IDs 
(e.g., $r=500$, $h=4$)
of top-$h$ most similar landmarks 
$
\{(v_1, \cdots, v_h)| d(x, p_{v_1})<\cdots<d(x, p_{v_h})\}, 
$
which can then be efficiently indexed using an inverted index. 
(2) VPR-to-NBV knowledge transfer: Second, we address the challenge of RL under uncertainty (i.e., active self-localization) 
by transferring the state recognition ability of VPR to the NBV. 
Our scheme is based on a recently developed method of reciprocal rank feature (RRF) \cite{icra21takeda}, 
inspired by our previous works \cite{icra21takeda, kanji2019detection, kanji2015unsupervised},
which is effective for transfer learning 
\cite{distil}
and information retrieval 
\cite{mmir}. 
(3) NNQL-based NBV: Third, we regard the available VPR as the experience database by adapting nearest-neighbor approximation of Q-learning (NNQL) \cite{nnql}. 
The result is an extremely compact data structure that compresses the VPR and NBV into a single incremental inverted index. Experiments using the public NCLT dataset validated the effectiveness of the proposed approach.

\arxiv{}{

\section{%
Related work
}

Self-localization aims to estimate the robot's viewpoint with respect to a pre-constructed model (``map") of the robot workspace. 
Various types of map models with different characteristics have been developed 
\cite{IBOW, topo2, neira2003linear}. 

A most popular approach is to model a map as an image retrieval system, in which the map images are indexed with their vectorial visual features. In this context, efforts pertaining to visual features (e.g., global features 
\cite{topo2}, 
local features \cite{lowe2004distinctive}), 
indexing (e.g., bag-of-words \cite{topo1}, incremental indexing \cite{IBOW}),
inference (e.g., pose tracking 
\cite{MCL}, map matching \cite{neira2003linear}), 
and post-verification (e.g., geometric verification \cite{matas2004randomized}). 
However, their per-domain retraining cost under domain shifts (e.g., 
change detection 
\cite{changedetection}, map updating \cite{update}) 
may be enormous because these retrieval systems exhaustively maintain the map of the entire workspace. 

An alternative approach is to model a map as an image classifier \cite{ref1} that regards a live view image as a query and predicts its place class. This model is based on the availability of predefined place classes, which are typically obtained by partitioning the robot workspace into place regions. Their per-domain cost for retraining might be low owing to the domain adaptation ability of recent deep classifier learning techniques \cite{resnet}. 

Another alternative approach is to model the map as (dis)similarities to main landmark objects that represent the robot workspace \cite{burgardlandmark}. The per-domain cost for retraining is inherently low because the map encompasses only a small fraction (e.g., 2\%) of the robot workspace (i.e., the landmark regions). This landmark-based map model is the main focus of our study.

Active self-localization has been investigated in several contexts, where in-domain settings are primarily assumed \cite{burgard1997active,feder1999adaptive,stachniss2004exploration,hsiao2020aras,lehner2017exploration,chaplot2018active,deepactivelocalization,chaplot2020learning}. 
In 
\cite{burgard1997active}, an active self-localization task was addressed using an entropy-based viewpoint planner characterized as a mixture of Gaussians and by extending the Markov localization framework for action planning. 
A trade-off between exploration efficiency and map quality is considered in the context of active SLAM 
\cite{stachniss2004exploration,hsiao2020aras,lehner2017exploration},
while our study focuses purely on the VPR task.
In 
\cite{chaplot2018active}, a deep neural network-based extension of active self-localization was addressed using a learned policy model. In 
\cite{deepactivelocalization}, an advanced deep neural network was presented, in which not only the policy model, but also the perceptual and likelihood models are completely learned. In 
\cite{chaplot2020learning}, a neural-network-based active SLAM framework was investigated.

These existing studies assumed an in-domain scenario (e.g., indoor monotonous-textured maze-like environments \cite{chaplot2018active}), 
where appearance variations were small. This is unlike our cross-domain self-localization scenarios, where the appearance of scenes can be affected significantly by geometric conditions (e.g., viewpoint drifts, car parking, building construction), as well as by photometric conditions (e.g., time of day, weather, and season). 
Hence, we propose to transfer the domain-invariant (or domain-adaptive) state recognition ability of recent VPR techniques (e.g., deep neural networks) to the NBV planner; however, this method increases the complexity during NBV training.

SIMBAD has recently garnered attention for use in many perception applications where traditional vectorial feature descriptors are difficult to obtain or inefficient for learning. In \cite{an2015person}, 
a reference-based descriptor was introduced to address the problem of person reidentification, in which the descriptor for a person is translated from the original color or texture descriptors to similarity measures between the person and exemplars in the reference set. In \cite{cao2019random}, 
dissimilarity representation was applied to multiview classification tasks by establishing dissimilarity representations for each view and averaging these dissimilarities over the views. In \cite{mendiola2017bio}, 
a comparative study among several approaches using entropy template selection (ETS) and dissimilarity representations (DRs) was conducted to investigate the application of biochemical substances, where a combination of ETS and DRs outperformed the baselines.

In our study, the use of SIMBAD was investigated not only for use in perception (i.e., VPR), but also for action planning (i.e., NBV). In our opinion, landmark-based self-localization is an application where vectorial feature descriptions are 
inefficient,
because 
their space cost is proportional to the number and dimensionality of map features.
In our approach,
a scene is compactly described as an ordered list of 
$h (\ll r)$ 
most similar prototypes (e.g., $r=500$, $h=4$).
The landmark set requires 
very small space,
because 
they encompass only a small fraction 
(e.g., 2\%) 
of the robot workspace. 
Furthermore, VPR and NBV exhibit the same indexing structure, which renders our system extremely compact and scalable. To the best of our knowledge, active landmark-based self-localization in this context has not been investigated hitherto.

}

\noeditage{
\figC
}

\section{%
Approach
}

The active self-localization system
aims to estimate the robot location
on a prelearned route (Fig. \ref{fig:C}).
It iterates
for each viewpoint,
three main steps:
scene description,
(passive) self-localization,
and
NBV planning.
The scene description describes an input map/live image $x$ as its dissimilarities $D(x, \cdot)$ from the predefined landmarks. 
The (passive) self-localization incorporates each perceptual/action measurement into the belief of the robot viewpoint. 
The NBV planning takes the scene descriptor $x$ of a live image
and 
determines the NBV action.
In addition, three additional modules exist, 
i.e., landmark selection, mapping, and nearest-neighbor Q-learning, which operates offline to learn the prior knowledge 
(``landmarks," ``map," and ``Q-function") required for the abovementioned three main modules. These individual modules are described in detail in the following subsections.

\subsection{%
Landmark Model
} \label{sec:landmark}

The proposed landmark model can be explained as follows. Let $R=\{p_1, p_2, \cdots, p_r\}$ be a set of $r$ 
predefined prototypes (i.e., landmarks).
For a dissimilarity measure $d$, which measures 
the dissimilarity between a prototype $p$ and an input object $x$, one may consider a new description based on the proximities to the set $R$, as $D(x,R)=$$[d(x,p_1)$, $d(x,p_2)$, $\cdots$, $d(x,p_r)]$. Here, $D(x,R)$ is a 
data-dependent mapping $D(\cdot, R)$: $X$ $\rightarrow$ $\mathbb{R}^r$ from a representation $X$ to the 
dissimilarity space, defined by set $R$. This is a vector space, in which each dimension corresponds to a dissimilarity $D(\cdot, p_i)$ to the prototype from $R$. A vector $D(\cdot, p_i)$ of dissimilarities to the prototype $p_i$ can be interpreted as a feature. 
It is noteworthy that 
even when
no landmark is present in the input image,
the proposed model still can describe the image
as dissimilarities to the landmarks.

\noeditage{
\figA
}

The advantage of this representation is 
its applicability to 
any method in dissimilarity spaces,
including 
recent deep learning techniques.
For example,
$d$ can be 
(1)
a pairwise comparison network (e.g., deep Siamese \cite{zhan2017change}) that predicts 
the dissimilarity $d(x, p)$ between $p$ and $x$,
(2)
a deep anomaly detection network (e.g., deep autoencoder \cite{an2015variational}) that predicts the deviation $d(x|p)$ of $x$ from learned prototypes $p$,
(3)
an object proposal network (e.g., feature pyramid network \cite{kim2018parallel}) that predicts object bounding-boxes and class label $d(p|x)$ (e.g., \cite{shogo2013partslam}),
(4)
an L2 distance $|f(x)-f(p)|$ of deep feature $f(\cdot)$ 
\cite{babenko2014neural}
between $p$ and $x$,
as well as 
(5)
any methods for
dissimilarities 
$d(p, x)$
including
those for non-visual modalities (e.g., natural language).
In the current study,
the experimental system 
was based on
(4) 
using NetVLAD \cite{NetVLAD} as the feature extraction network $f(\cdot)$.
We believe that 
this 
enables effective exploitation of 
the discriminative power of a deep neural network within SIMBAD.

\subsection{%
Landmark Selection
}

The landmark selection is performed prior to the training/testing phase, in an independent ``landmark domain''.
Landmarks should be selected 
such that they are dissimilar to each other. This is because, if the dissimilarity $d(p_i, p_j)$ is 
small for a prototype pair, $d(x, p_i)\simeq d(x, p_j)$ for other objects $x$, 
then either $p_i$ or $p_j$ is a redundant prototype. Intuitively, 
an ideal clustering algorithm 
may identify good landmarks 
as cluster representatives. 
However, this clustering is NP-hard \cite{kanji2020self}. 
In practice, heuristics such as k-means algorithms are often used as alternatives.
Moreover, the utility issue 
(e.g., \cite{tanaka2003viewpoint})
complicates the problem,
i.e., landmark visibility and other observation conditions (e.g., occlusions and field-of-views) 
must be considered to 
identify useful landmarks. 
For such landmark selection, 
only a heuristic approximate solution exists,
i.e., no analytical solution exists \cite{topo3}. 

In our study, we do not focus on the landmark selection method; instead, we propose a simple and effective method. Our solution comprises three procedures: (1) First, high-dimensional vectorial features $X=\{x\}$ 
(i.e., NetVLAD) are extracted from individual candidate images. (2) Subsequently, each candidate image 
$x\in X$
is scored by the dissimilarity 
$\min_{x'\in X\setminus\{x\}} |x-x'|$ 
from its nearest neighbor over the other features. 
(3) Finally, all the candidate images are sorted in the descending order of the scores, and the top-$r$ images are selected as the prototype landmarks. This approach offers two advantages. First, the selected landmarks are expected to be dissimilar to each other. In addition, the dissimilarity features $D(\cdot, R)$ are expected to become accurate 
when the robot approaches a viewpoint with landmark view.

\subsection{%
Self-localization
}\label{sec:sl}

The (offline) mapping is performed prior to the self-localization tasks,
in an independent ``training domain''.
It extracts a 4,096 dim NetVLAD image feature from each map image $x$  
and then translates it to an $r$-dim dissimilarity descriptor 
$D(x, \cdot)$. 
Subsequently, it sorts the 
$r$ 
elements 
of the descriptor in the ascending order and returns an ordered list of top-$h$ ranked landmark IDs,
which is our proposed $h$-dim scene descriptor (Fig. \ref{fig:A}).
With an inverted index,
the map image's ID 
is 
indexed 
by using its landmark ID 
and the rank value,
as primary and secondary keys (Fig. \ref{fig:B}).

The (online) self-localization
is performed in a new ``test domain''.
It translates an input query image 
to a ranking-based descriptor. Subsequently, the inverted index is looked up using each 
landmark ID in the ordered list as a query. Hence, a short list of map images with a common 
landmark ID is obtained. Next, the relevance of a map image is evaluated as the inner product
$\langle v_{RRF}^r(q), v_{RRF}^h(m) \rangle$ 
of RRF
between the query image $q$ and each map image $m$ in the short list. 
An RRF $v_{RRF}^h$ is an $r$-dim $h$-hot vector whose 
$i$-th element is the reciprocal 
$1/v[i]$ 
of the rank value $v[i]$ of the $i$-th landmark if 
$v[i]\le h$,
or 0 otherwise. For more details regarding the RRF, please refer to \cite{icra21takeda}.

During the active multi-view self-localization,
the results of (passive) self-localization at each viewpoint
are
incrementally integrated 
by a particle filter (PF).
PF is
a computationally efficient implementation of a Bayes filter, 
and can address multimodal belief distributions \cite{MCL}.  
In the PF-based inference system,
the robot's 1D location on the route
prelearned in the training domain
is regarded as the state.
The PF system is initialized at the starting viewpoint of the robot. In the prediction stage, the particles are propagated through a dynamic model using the latest action of the robot. In the update stage, the likelihood models for the landmark observation are approximated by the RRF, and the weight of each $k$-th particle is updated as
$w_k$$\leftarrow$$w_k$+$v_{RRF}[u_k]$,
where $v_{RRF}[u_k]$ is the RRF element that corresponds to the hypothesized viewpoint $u_k$. 
Other details are the same as that of the PF-based self-localization framework \cite{MCL}.

\noeditage{
\figB
}

\subsection{%
NBV Planning
}

The NBV planning task is formulated as an RL problem, 
in which a learning agent interacts with a stochastic environment. 
The interaction is modeled as a discrete-time discounted Markov decision process (MDP). 
A discounted MDP is a quintuple 
$(S, A, P, R, \gamma)$, 
where $S$ and $A$ are the set of states and actions, 
$P$ the state transition distribution, 
$R$ the reward function, and 
$\gamma \in(0,1)$ a discount factor ($\gamma=0.9$). 
We denote by 
$P(\cdot|s, a)$ and $R(s, a)$ 
the probability distribution over the next state and the immediate reward of performing action $a$ at state $s$, respectively.

A Markov policy  is the distribution over the control actions for the state, 
in our case represented by the scene descriptor of a live image (i.e., $s=x$).
The action-value function of a policy $\pi$, 
denoted by $Q:$ $S\times A$ $\rightarrow$ $\mathbb{R}$,
is defined as the expected sum of discounted rewards that are encountered when 
policy
$\pi$
is executed. 
For an MDP, the goal is to identify a policy that yields the best possible values,
$Q^*(s, a)$ = $\sup_\pi$ $Q^\pi(s, a)$, $\forall(s, a)\in S\times A$.

The implementation of the ``Q-function" $Q(\cdot, \cdot)$ as a computationally tractable function is a key issue. 
A naive implementation of the function $Q(\cdot, \cdot)$ is 
to employ a two-dimensional table, which is indexed by a state-action pair $(s, a)$ and whose element is a Q-value \cite{sutton_org}. 
However, this simple implementation presents several limitations. 
In particular, it requires a significant amount of memory space, 
proportional to the number and dimensionality of the state vectors, 
which are intractable in many applications including ours. 
An alternative approach 
is to use a deep neural network based approximation of the Q-table (e.g., DQN \cite{dqn}),
as in \cite{9512691}. 
However, 
the
DQN
must be retrained
for every new domain
with
a significant amount of space/time overhead.
Herein,
we present a new highly efficient NNQL-based approximation
that reuses the existing inverted index (\ref{sec:sl}) to approximate $Q(\cdot,\cdot)$ 
with a small additional space cost. 

Our approach is inspired by the recently
developed nearest-neighbor approximation of the Q-function (Fig. \ref{fig:B}) 
\cite{nnql}. 
Specifically, the inverted index 
which was originally developed for VPR (\ref{sec:sl}) 
is regarded as a compressed representation of visual experiences in the training domain. 
Recall that the inverted index aims to index map image ID using the 
landmark ID. 
Next, we now introduce a supplementary two-dimensional table
called ``experience database"
that aims to index action-specific Q-values $Q(s, \cdot)$ by map image ID. 
Subsequently,
we 
approximate the Q-function by the NNQL \cite{nnql}. 
The key difference 
between 
NNQL
and the standard Q-learning is that
the Q-value of
an input state-action pair $(s, a)$
in the former
is approximated
by a set of Q-values that are associated
with its $k$ nearest neighbors ($k=4$).
Next,
we used
the supplementary table to store the action-specific Q values $Q(s, \cdot)$. 
Hence, the Q-function is approximated by
$|N(s|a)|^{-1} \sum_{(s', a) \in N(s|a)} Q(s', a)$,
where
$N(s|a)$ is the nearest neighbor of $(s, a)$ conditioned on a specified action $a$.
In fact, such an action-specific NNQL can be regarded as an instance of RL. For more details regarding NNQL, please refer to \cite{nnql}.

\noeditage{
\figH
}

\section{Experiements}

We evaluated the effectiveness of the proposed algorithm via self-localization tasks in a cross-season scenario. 

We used the publicly available NCLT dataset \cite{NCLT}. 
The data we used included view image sequences along vehicle trajectories acquired using the front-facing camera of the Ladybug3, as well as ground-truth GPS viewpoint information. Both indoor and outdoor change objects such as cars, pedestrians, construction machines, posters, and furniture were present during seamless indoor and outdoor navigation by the Segway robot. 

In the experiments, we used four different datasets, i.e., 
``2012/1/22 (WI)", ``2012/3/31 (SP)", ``2012/8/4 (SU)", and ``2012/11/17 (AU)",
which contained 
26,208, 26,364, 24,138, and 26,923 images (Fig. \ref{fig:H}). 
They were used to create eight different landmark-training-test domain triplets (Fig. \ref{fig:C}): 
WI-SP-SU, 
WI-SP-AU, 
SP-SU-AU,
SP-SU-WI,
SU-AU-WI,
SU-AU-SP,
AU-WI-SP,
and
AU-WI-SU.
The number of landmarks was set to 500 for every triplet, which is consistent with the settings in our previous studies using the same dataset 
(e.g., \cite{ref15}).

For NNQL training, 
the learning rate was set $\alpha=0.1$.
Q-value for a state-action pair was initialized to 0.0001,
and a positive reward of 100 was assigned when the belief value of the ground-truth viewpoint was top-10\% ranked.
The number of training episodes was 10,000.
The action candidates are a set of forward movements $A=\{1, 2, \cdots, 10\}$ [m] 
along the route
defined in the NCLT dataset. 
One episode consists of an action sequence of length 10, 
and its starting location is randomly sampled from the route.

\noeditage{

\arxiv{}{\figE}

\figFJK 

\figG

}

The VPR performance at each viewpoint in the context above was measured based on the averaged normalized rank (ANR). In the ANR, a subjective VPR is modeled as 
a ranking function that takes an input query image and assigns a rank value to each map image. Subsequently, the rank values for the ground-truth map images are computed, normalized by the number of map images, and averaged over all test queries, thereby yielding the ANR. 
A VPR system with high ANR performance can be regarded as 
having nearly perfect performance 
and high retrieval robustness \cite{RetrievalRobustness}, 
which is typically required in loop closing applications \cite{joan2020lipo}.

\arxiv{}{
Figure \ref{fig:E} shows examples of landmarks used for ranking-based scene descriptions. As shown, each input image is described by spatially distant but similar landmark images, owing to the descriptive power of the deep feature extraction network.
}

Four different scene descriptors based on NetVLAD were developed as the baseline/ablation methods: ``brute-force", ``SIMBAD", 
``bag-of-(most-similar-)landmarks", and ``RRF". In ``brute-force", the map model is a nearest-neighbor search based on the L2 dissimilarity measure with NetVLAD of the query image, assuming that every map image is described by the NetVLAD feature. In the other three methods, the map model describes each map image 
by the dissimilarities with the $h$ most similar landmarks. More specifically, 
``bag-of-landmarks" assumes $h$-hot binary (0/1) similarity value, 
``SIMBAD" uses the $h$-hot L2 dissimilarity value, and 
``RRF" uses the $h$-hot RRF similarity value. 

It is noteworthy that the three methods 
``SIMBAD", ``bag-of-landmarks", and ``RRF"
require significantly lower time cost owing to the availability of efficient inverted index. 

In terms of space cost, ``bag-of-landmarks" is the most efficient, ``RRF" and ``SIMBAD" are slightly more expensive because they need to memorize the rank value for each map image, and ``brute-force" is intractably expensive as it requires to memorize the high-dimensional map features.

Figure \ref{fig:FJK} shows ANR of single-view VPR tasks.

As can be seen from Fig. \ref{fig:FJK} (a), the ``brute-force" method shows very good recognition performance, but at the cost of high time/space cost for per-domain retraining. As we found in \cite{icra21takeda}, NetVLAD's utility as a dissimilarity-based feature vector is low. In this experiment, the method ``SIMBAD" was about the same as ``bag-of-landmarks" and had slightly lower performance. Compared with these two methods, the proposed method ``RRF" had much higher performance. 
To summarize, the proposed method ``RRF" achieves a good trade-off between 
time/space efficiency and recognition performance. In subsequent experiments, we will use this ``RRF" as the default method for the passive self-localization module. 

\arxiv{}{

As can be seen from Fig. \ref{fig:FJK} (b), the relative superiority of the proposed method over the other two methods becomes remarkable when the number of landmarks is small (e.g., $r=10$). Accuray of the proposed method increases monotonically as the number of landmarks increases. 

}

As can be seen from Fig. \ref{fig:FJK} 
\arxiv{(b), }{(c), }
in the case where the rank list length (i.e., descriptor size) is very small (e.g., $h=4$), the proposed method has a small performance drop, while the other two methods have as bad VPR performance as the chance method (i.e., ANR=50\%).

Figure \ref{fig:G} 
demonstrates 
the 
performance of 
active self-localization.
Note that each episode has a different number of valid observations, and long episodes do not always provide high performance. We plot the ANR performance for all test images (frames) on the graph in Fig. \ref{fig:G}, 
regardless of how many episodes were observed. In the figure, ``proposed" is the performance of NBV using NNQL trained 
using the RRF feature as a state vector. ``baseline" is different from ``proposed" only in that the action is randomly determined. 
``$a$[m]" ($a\in[1, 10]$)
is a naive strategy that
repeats the same forward movement $a[m]$ at every viewpoint,
independent of the RRFs. 
They often performed worse than the ``proposed",
and to make the matters worse,
which action is best is domain-dependent and cannot be known in advance.
The characteristics 
that the higher the VPR performance, the better the active self-localization performance, 
is consistent with our recent studies
 dealing with other VPR methods
(e..g, pole-landmark-based VPR \cite{9512691},
e.g., convolutional neural network -based VPR \cite{9412043},
and
e.g., map-matching-based VPR \cite{9493670}).
It is noteworthy that 
the proposed method outperformed all the methods considered here, 
although the landmark-based VPR and NNQL-based NBV were compressed into a single incremental inverted index.

Finally, we also investigated the space/time performance.
The inverted index consumes 15$h$-bit per map image.
The main processes,
extracting NetVLAD features,
VPR,
and
NBV
consumes
8.6 ms (GPU),
495.4 ms (CPU),
and
7$\times 10^{-3}$ ms (CPU)
per viewpoint
(CPU: intel core i3-1115G4 3.00GHz).
The most time consuming part 
is the scene description processing
in VPR,
which consumes
410 ms.
The NBV is significantly fast,
compared with the recent deep learning variants,
thus
the additional cost for
the extension from the passive to active self-localization
was little.

\section{Conclusions and Future Directions}

Herein, a new framework for active landmark-based self-localization
for highly compressive applications
(e.g., 40-bit image descriptor)
was proposed.
Unlike previous approaches of active self-localization, we assumed the main landmark objects as the only available model of the robot's workspace (``map"), and proposed to describe each map/live scene as dissimilarities to these landmarks. Subsequently, a novel reciprocal rank feature-based VPR-to-NBV knowledge transfer was introduced to address the challenge of RL under uncertainty. Furthermore, an extremely compact data structure that compresses the VPR and NBV into a single incremental inverted index was presented. The proposed framework was experimentally verified using the public NCLT dataset.

Future work must investigate how to accelerate the (dis)similarity evaluator $d(\cdot, \cdot)$ while retaining the ability to discriminate between different levels of (dis)similarities. In the training phase, the evaluator is repeated for every viewpoint of every episode. Since the number of repetitions is very large (e.g., 
10,000$\times$10$\times$500=5$\times$$10^7$), efficiency of calculation is the key to suppress per-domain retraining cost.
Another direction for future research is compression of the proposed scene/state descriptor, towards extremely-compressive applications (e.g., \cite{yan2019global}). Although the proposed ranking-based descriptor is the first compact (e.g., 40-bit) descriptor for the VPR-to-NBV applications, it is still uncompressed, i.e., it may be further compressed. Finally, our on-going research topic is to incorporate various (dis)similarity evaluators into the proposed deep SIMBAD framework, including non-visual -based (dis)similarity evaluators (\ref{sec:landmark}).

\arxiv{
}{
\newpage
}

\editage{

\newpage ~\\

\figD

\figC

\figA

\figB

\figH

\arxiv{}{
\figE
}

\figFJK 

\figG


\newpage ~
\newpage ~
\newpage ~

}

\bibliographystyle{IEEEtran}
\bibliography{../ds}

\end{document}